\documentclass{article}

\usepackage{arxiv}
\usepackage{siunitx}
\usepackage{booktabs}
\usepackage{multirow}
\usepackage{algorithm}
\usepackage{algorithmic}
\usepackage{dsfont}
\usepackage{caption}
\usepackage{tabularx}
\usepackage[export]{adjustbox}
\usepackage{tikz}
\usepackage{amssymb}
\usepackage{amsmath}
\usepackage{hyperref}
\DeclareMathOperator*{\argmax}{arg\,max}
\DeclareMathOperator*{\argmin}{arg\,min}

\newtheorem{theorem}{Theorem}[section]

\newtheorem{Proposition}[theorem]{Proposition}

\newtheorem{Remark}[theorem]{Remark}
\newtheorem{Proof}[theorem]{Proof}
\newtheorem{Theorem}[theorem]{Theorem}

\title{Learning from both experts and data.}

\author{
  R\'emi~Besson \\
  CMAP\\
  \'Ecole Polytechnique\\
  Route de Saclay, 91128 Palaiseau \\
  \texttt{remi.besson@polytechnique.edu} \\
   \And
 Erwan~Le Pennec \\
   CMAP\\
  \'Ecole Polytechnique\\
  Route de Saclay, 91128 Palaiseau \\
  \texttt{erwan.le-pennec@polytechnique.edu} \\
  \And
 St\'ephanie~Allassonni\`ere \\
  School of Medicine\\
  Paris-Descartes University\\
  15 Rue de l'\'Ecole de M\'edecine, 75006 Paris \\
  \texttt{stephanie.allassonniere@parisdescartes.fr} \\
}

\begin{document}
\maketitle

\begin{abstract}
In this work we study the problem of inferring a discrete probability distribution using both expert knowledge and empirical data. This is an important issue for many applications where the scarcity of data prevents a purely empirical approach. In this context, it is common to rely first on an initial domain knowledge a priori before proceeding to an online data acquisition. We are particularly interested in the intermediate regime where we do not have enough data to do without the initial expert a priori of the experts, but enough to correct it if necessary. We present here a novel way to tackle this issue with a method providing an objective way to choose the weight to be given to experts compared to data. We show, both empirically and theoretically, that our proposed estimator is always more efficient than the best of the two models (expert or data) within a constant.
\end{abstract}

\newcommand{\enstq}[2]{\left\{#1~\middle|~#2\right\}}

\section{Introduction}
In this work we present a novel way to estimate a discrete probability distribution, denoted $p^\star$, using both expert knowledge and data. This is a crucial aspect for many applications. Indeed when deploying a decision support tool we often rely entirely on expert/domain knowledge at the beginning and the data only then comes with the use of the algorithm in real life. However, we need a good model of the environment directly to train the decision support tool with a planning algorithm. This model of the environment is to be refined and corrected as the data flow increases.

We assume here to have some expert knowledge under the form of an initial a priori on the marginals, the moments and/or the support of $p^\star$ or any other relevant information. We also assume that we sequentially receive data. We denote $x^{(1)}$,..., $x^{(n)}$  an i.i.d sample from $p^\star$.

One example of application may come from the objective of building a symptom checker for rare diseases \cite{besson2018}. In this case, $p^\star$ represents the probability of the different possible combinations of symptoms given the event that the disease of the patient is $D$. More precisely, we denote:

\begin{equation}
    p^\star=(p_1^\star,...,p_K^\star)^T=\left(\begin{matrix} \mathbb{P}[\bar{B_1},...,\bar{B}_{J-1},\bar{B_J}\mid D]\\\mathbb{P}[\bar{B_1},...,\bar{B}_{J-1},B_J\mid D]\\ \vdots \\ \mathbb{P}[B_1,...,B_{J-1},B_J\mid D] \end{matrix}\right)
\end{equation}

the distribution we aim to estimate where $D$ is the random variable disease. $B_1$,..., $B_J$ are the typical symptoms of the disease $D$, all are binary random variable i.e the symptom can be present or absent. We aim to estimate the $2^J=K$ different combinations (as $\mathbb{P}[B_1,...,B_L\mid D]$ for example) when we only have an expert a priori on the marginals $\mathbb{P}[B_i\mid D]$, for all $i\in[1,J]$. 

Of course a first idea would be to assume that the symptoms are conditionally independent given the disease. However, we expect complex correlations between the typical symptoms of a given disease. Indeed we can imagine two symptoms very plausible individually but which rarely occur together (or even never in the case of incompatible symptoms like for example microcephaly and macrocephaly).

Note also that the  assumption of conditional independence would make it possible to present a disease without having any of the symptoms related to this disease in the database (when there is no $B_i$ such that $\mathbb{P}[B_i\mid D]=1$), which should be impossible.

Generally speaking if we had enough empirical data, we would no longer need the experts. Conversely, without empirical data, our model must be based entirely on experts. We detail here two different approaches to deal with intermediate regime where we do not have enough data to do without the a priori given by the experts but where we have enough data to correct and specify this initial a priori. This approaches are meaningful as long as we do not know how much data have been used to build the initial a priori and then that we really try to combine two heterogeneous form of information : experts and empirical data.   

We first recall in section \ref{maxenttheory} the principle of maximum entropy which is the basic brick we use to build an expert model. We then briefly introduce the proposed approach to mix expert and data in section \ref{barycenterexp}. We underline to what extent this approach is superior to the one we previously proposed in \cite{besson2018}.
The barycenter approach we propose here provides an objective way to choose the weight to be given to experts compared to data. On the contrary, the maximum-likelihood with entropic penalization approach of \cite{besson2018} was shown to be sensitive to the choice of the regularization parameter. We make in section \ref{litteratureexperdata} a review of the literature. We finally show in section \ref{barycentertheory}, both empirically and theoretically, that our barycenter estimator is always more efficient than the best of the two models (expert or data) within a constant.

Note that even if we will refer throughout the paper to our particular application in medicine our framework is relevant for any inference problem involving an initial a priori with a particular form (marginals, moments, support,...) combined with data. Biology, ecology and physics, to name a few, are areas where ideas of maximum entropy have been used for a long time and then where the ideas developed in this work could be interesting. See \cite{martino} for an overview of the maximum entropy applications for inference in biology.

\section{Mixing expert and empirical data}

\subsection{Building an expert model: the maximum entropy principle}\label{maxenttheory}

The aim to benefit simultaneously from expert data and empirical data has of course a very old history. This is the very essence of Bayesian statistics \cite{gelmanbda04} which aims to integrate expert data, in the form of an a priori, which is updated with empirical data using the Bayes' theorem to obtain what will be called the posterior.

Note that in our case we do not have a classical a priori modeling the model parameters with probability distributions. We have an a priori on the marginals as such as a number of constraints on the distribution to be estimated. The absence of an obvious a priori to model the distribution of the parameters naturally leads us to the idea of maximum entropy theorized by \cite{JaynesInformationTA}. Indeed, if no model seems more plausible to us than another, then we will choose the least informative. This is a generalization of the principle of indifference often attributed to Laplace:

"We consider two events as equally probable, when we see no reason that makes one more probable than the other, because, even if there is an unequal possibility between them, since we don't know which is the biggest, this uncertainty makes us look at one as as likely as the other" \cite{laplace1774}.

This principle therefore takes the form of an axiom allowing us to construct a method to choose an a priori: the least informative possible consistent with what we know. 

We then define the distribution of maximum entropy as follow:

\begin{equation}\label{maxent}
p^{\text{maxent}}=\argmax_{p/ p\in \tilde{\mathcal{C}}} H(p)
\end{equation} 

where $\tilde{\mathcal{C}}=\mathcal{C}\bigcap \mathcal{C}^{\text{expert}}$. $\mathcal{C}=\{p/\sum_i p_i=1, p_i\geq 0\}$ is the probability simplex and $\mathcal{C}^{\text{expert}}$ is the set of constraints fixed by experts. 

Note that $p^{\text{maxent}}$ is well-defined, namely it exists and is unique, as long as $\mathcal{C}^{\text{expert}}$ is a convex set. Indeed the function $p\mapsto H(p)$ is strictly concave and it is a classic result that a strictly concave function under convex constraints admit an unique maximum.  

It is well-known that if $\mathcal{C}^{\text{expert}}$ only contained the constraints for the marginals then $p^{\text{maxent}}$ is nothing more that the independent distribution.

However, in our case, we can add some information about the structure of the desired distribution as constraints integrated to $\mathcal{C}^{\text{expert}}$. We judge impossible to have a disease without having at least a certain amount of its associated symptoms: one, two or more depending on the disease. Indeed the disease we are interested in manifest themselves in combination of symptoms. The combinations allowing the fact to have simultaneously two exclusive symptoms should also be constraints to be equal to $0$. All combinations of constraints are conceivable as long as $\tilde{\mathcal{C}}$ remains a convex closed space, in order to ensure the existence and uniqueness of $p^{\text{maxent}}$.

We therefore construct our a priori by taking the maximum entropy distribution checking the constraints imposed by the experts. Thus among the infinite distributions that verify the constraints imposed by the experts, we choose the least informative distribution $p^{\text{maxent}}$, in other words the one closest to the conditional independence distribution.

We need to add information to move from the information provided by the experts to the final distribution and we want to add as little as possible on what we don't know. This approach is referred to as maxent (maximum entropy) and has been widely studied in the literature \cite{JaynesInformationTA}, \cite{Cover:2006:EIT:1146355}, \cite{Berger:1996:MEA:234285.234289}.

\subsection{Barycenters between experts and data }
\label{barycenterexp}

Recall that $p^\star=(p_1^\star,...,p_K^\star)$ and that $x^{(1)}$,..., $x^{(n)}$ is an i.i.d sample of $p^\star$. The empirical distribution $p_n^{\text{emp}}=(p_{n,i}^{\text{emp}})_{i=1}^{K}$ is given by:

\begin{equation}\label{emp}
p_{n,i}^{\text{emp}}=\displaystyle\frac{1}{n} \sum_{j=1}^{n} \mathds{1}_{\{x^{(j)}=i\}}.
\end{equation}

Following the ideas of section \ref{maxenttheory} we define the expert distribution as the distribution which maximize entropy while satisfying the constraints fixed by experts :

\begin{equation}\label{expertmaxent}
p^{\text{expert}}=\argmax_{p/ p\in \tilde{\mathcal{C}}} H(p)
\end{equation} 

where $\tilde{\mathcal{C}}$ is the intersection of the simplex probabilities with the set of constraints fixed by experts: in our case it is composed of a list of censured combinations and a list of marginals given by experts. Note that it is possible to give more or less credit to the marginals given by experts by formulating the constraint as an interval (more or less wide) rather than a strict equality. The distribution of expert is then defined as the least informative distribution consistent with what we know. 

Let $\mathcal{L}$ be any dissimilarity measure between two probability distributions. Our barycenter estimator mixing expert and empirical data is then defined as: 
\begin{equation}
\label{formulegenerale}
    \hat{p}_{\epsilon_n}^{\mathcal{L}}=\argmin_{p \in \mathcal{C}/ \mathcal{L}(p_n^{\text{emp}},p) \leq \epsilon_n } \mathcal{L}(p^{\text{expert}},p)
\end{equation}

where  
\begin{equation}
\epsilon_n:=\epsilon_n^{\delta}=\argmin_{l} \mathbb{P}[ \mathcal{L}(p_n^{\text{emp}},p^{\star}) \leq l]\geq 1-\delta.
\end{equation}

$\hat{p}_n^{\mathcal{L}}$ is then defined as the closest distribution from experts, in the sense of the dissimilarity measure $\mathcal{L}$, which is consistant with the observed data.  

For such a construction to be possible, we will therefore have to choose a measure of dissimilarity $\mathcal{L}$ such that we have a concentration of the empirical distribution around the true distribution for $\mathcal{L}$. 

Such a formulation has several advantages over the maximum likelihood with entropic penalization approach previously proposed in \cite{besson2018}. First, we do not have to choose a regularization parameter which seems to have a strong impact on the results of the estimator (see \cite{besson2018}). This parameter is replaced by the parameter $\delta$, that it is reasonable not to take more than $0.1$ and which appears to have low impact on the result of $\hat{p}_n^{\mathcal{L}}$ (see section \ref{numericBarycenter}). Secondly the solution of \eqref{formulegenerale} can be (it of course depends on the choice of the dissimilarity measure $\mathcal{L}$) easier to compute than the one of the optimization problem associated to the penalization approach for which a closed form of the solution could not be derived \cite{besson2018}.

\section{Related works}
\label{litteratureexperdata}

\subsection{Bayesian statistics}
The desire to take advantage of expert data and empirical data at the same time has of course a very old history. This is the very essence of Bayesian statistics \cite{gelmanbda04} which consists in incorporating expert data, in the form of an a priori, into experimental data by using Bayes' theorem to obtain what will be called the posterior. 

Our a priori concerns the marginals and a certain number of constraints on the distribution to be estimated. The absence of an obvious a priori to modelize the parameter's distribution naturally leads us to the idea of maximum entropy theorized by \cite{JaynesInformationTA}. 

Indeed, if no model seems more plausible to us than another, then our choice will be the least informative. This is a generalization of the principle of indifference often attributed to Laplace: "We look at two events as equally probable, when we see no reason that makes one more probable than the other, because, even if there is an unequal possibility between them, as we do not know on which side is the greatest, this uncertainty makes us look at one as equally probable as the other" \cite{laplace1774}. This principle therefore takes the form of an axiom that allows us to construct a method for choosing an a priori: the least informative possible compatible with what we know.

\subsection{Expert system with probabilistic reasoning}

The creation of a decision support tool for medical diagnosis has been an objective since the beginning of the computer age. Most of the early work proposed a rules-based expert system, but in the 1980s, a significant part of the community studied the possibility of building an expert system using probabilistic reasoning \cite{Pearl1989ProbabilisticRI}. Bayesian probabilities and methods were therefore relatively early considered as good ways to model the uncertainty inherent in medical diagnosis. 

The assumption of conditional independence of symptoms given the disease has been intensively discussed as it is of crucial importance for computational complexity. Some researchers considered this hypothesis harmless \cite{Charniak1983TheBB} while others already proposed a maximum entropy approach to face this issue \cite{Hunter:1985:URU:3023810.3023813}, \cite{DBLP:journals/corr/abs-1304-3423} or \cite{DBLP:journals/corr/abs-1304-1104}.

However, it seems that none of the work of that time considered the expert vs empirical data trade-off that we face. In the review article \cite{DBLP:journals/kbs/Jirousek90} presenting the state-of-the-art of the research of that time (1990) about this issue, it is clearly mentioned that these methods only deal with data of probabilistic form. More precisely, they assume that they have an a priori on the marginal but also on some of the combinations of symptoms (in our case we would assume that we have a priori on $\mathbb{P}[B_1,B_2\mid D]$ for example) and propose a maximum entropy approach where these expert data are treated as constraints in the optimization process. Once again, this is not the case for us since we have only an a priori on the marginal (and a certain number of constraints) as well as experimental data. This field of research was very active in the 1980s and then gradually disappeared, probably due to the computational intractability of the algorithms proposed for the computer resources of the time. 

\subsection{Bayesian Networks} 
 Bayesian networks were then quickly considered as a promising alternative to model probabilistic dependency relationships between symptoms and diseases \cite{Pearl1989ProbabilisticRI}. These are now used in most expert systems, particularly in medicine \cite{Koller:2009:PGM:1795555}. 

A Bayesian network is generally defined as an acyclically oriented graph. The nodes in this graph correspond to the random variables: symptoms or diseases in our case. The edges link two correlated random variables by integrating the information of the conditional law of the son node given the father node. The main advantage of such a model is that it can factorize the joint distribution using the so-called global Markov property. The joint law can indeed be expressed as the product of the conditional distributions of each node given its direct parents in the graph \cite{spiegelhalter1993}.     

The construction of a Bayesian network implies first of all to infer its structure, i.e. to determine the nodes that must be linked by an edge of those that can be considered conditionally independent to the rest of the graph (structure learning). Then, learning the network implies learning the parameters, i.e. the probabilities linking the nodes (parameter learning).

It is therefore natural to also find in this area of the literature works that aimed at mixing expert and empirical data. In \cite{ZHOU201669} the experts' indications take a particular form since they indicate by hand correlations, positive or negative, between variables. The approach of \cite{Constantinou:2016:IEK:2927996.2928177} is also quite distant because it is preferably based on data. \cite{Constantinou:2016:IEK:2927996.2928177} only uses expert indications for additional variables for which there are no data, typically rare events never observed in the database. A work closer to ours is \cite{Heckerman1995} where the authors assume that they have a first Bayesian network built entirely by the experts, to which they associate a degree of trust. The authors then use the available data to correct this expert network. We distinguish ourselves from this work in our effort to find an objective procedure for the weight to be given to experts in relation to the data (and for this weight not to be set by the experts themselves).

Note also that the main interest of Bayesian networks is to take advantage of conditional independence relationships known in advance, as they are pre-filled by experts or inferred from a sufficient amount of data. However, in our case, we do not have such an a priori knowledge about the dependency relationships between symptoms and not enough data to infer them.  




\subsection{From the marginals to the joint distribution}

Estimating the joint distribution from the marginal is an old problem, which is obviously not necessarily related to expert systems. This problem is sometimes referred to in the literature as the "cell probabilities estimation problem in contingency table with fixed marginals". The book \cite{Bishop75discretemultivariate} gives a good overview of this field. We can trace back to the work of \cite{deming1940} which assumes knowing the marginal and having access to a sample of empirical data and aims to estimate the joint distribution. In this article, they proposed the "iterative proportional fitting procedure" (IPFP) algorithm, which is still very popular to solve this problem.

An important assumption of \cite{deming1940} is that each cell of the contingency table receives data. 
In \cite{Ireland1968ContingencyTW} the authors prove that the asymptotic estimator obtained by an IPFP algorithm is the distribution that minimizes the Kullback-Leibler divergence from the empirical distribution under the constraint to respect the marginal experts. 

However, an IPFP algorithm is not suitable for our problem for two main reasons: first, we do not have absolute confidence in the marginals given by experts (we want to allow us to modify them as we collect more data) and second, because since we are interested in rare diseases we do not expect to have a sufficient amount of data. In fact, many of the cells in the contingency table we are trying to estimate will not receive data, but it would be disastrous in our application to assign a zero probability to the corresponding symptom combination.     

In a sense, an IPFP algorithm does exactly the opposite of what we are aiming for: it modifies empirical data (as little as possible) to adapt them to experts, while we aim to modify experts (as little as possible) to make them consistent, in a less restrictive sense, with empirical data.

We should also mention the work related to our problem in applications of statistics to the social sciences where researchers aim to construct a synthetic population from marginal coming from several inconsistent sources \cite{Barthelemy2013SyntheticPG}. Their proposed approach also use ideas of maximum entropy but it is still different of our trade-off expert vs empirical data since they build their model without samples. 

\subsection{The Kullback centroid}

Our optimization problem \eqref{formulegenerale} in the particular case where the dissimilarity measure $\mathcal{L}$ is the Kullback Leibler divergence is called moment-projection (M-projection) in the literature. The properties of these projections have been intensely studied \cite{DBLP:journals/tit/CsiszarM03}.

Note that the Lagrangian associated with such an optimization problem is then nothing more than a Kullback-Leibler centroid. These objects or variations/generalization of them (with Jeffrey's, Bregman's divergences etc...) have been the subject of research since the paper of \cite{Veldhuis2002TheCO}. For example, articles
\cite{Nielsen2009SidedAS} and \cite{Nielsen2013JeffreysCA} study cases where an exact formula can be obtained and propose algorithms when this is not the case.

However, we have not found any use of these centroids to find a good trade-off expert vs empirical data as we propose in this paper. Bregman's divergence centroids have been used to mix several potentially contradictory experts, the interested reader may refer to the recent thesis of \cite{Adamcik2014}. We could certainly consider that the empirical distribution $p_n^{\text{emp}}$ is a second expert and that our problem is the same as to mix two experts: literature and data. However, the question of the weight to be given to each expert, which is the question that interests us here, will not be resolved. In \cite{Adamcik2014} the aim is rather to synthesize contradictory opinions of different experts by fixing in advance the weight to be given to each expert. We propose, for our part, an objective procedure to determine the weight to be given to experts comparing to empirical data.

\section{Numerical experiments and theoretical properties of the barycenter estimator}
\label{barycentertheory}
\subsection{Barycenter in normed spaces}
In this section we work in spaces $L^p$. Let us recall that the classic norm on the space $L^p$ is given by :  $\lVert x\rVert_j=\left(\displaystyle\sum_i |x_i|^j\right)^{\frac{1}{j}}$.

Following the ideas presented in section \ref{barycenterexp} we define our estimator, $\forall i \geq 1$, $\forall j \geq 1$ as follow :

\begin{equation}
    \hat{p}_n^{i,j}=\argmin_{p\in \mathcal{C}/ \lVert p- p^{\text{expert}} \rVert_i \leq \epsilon_n} \lVert p- p_n^{\text{emp}} \rVert_j 
    \label{optbarycenternormed}
\end{equation}

where 
\begin{equation}\label{confidence1}
\epsilon_n:=\epsilon_n^{\delta}=\argmin_{l} \mathbb{P}[\lVert p_n^{\text{emp}} -  p^{\star}\rVert_i \leq l]\geq 1-\delta.
\end{equation}

To control $\epsilon_n$ we use the concentration inequality obtained in the recent work of \cite{mardia2018concentration}. In the literature, most of the concentrations inequalities for the empirical distribution use the $L^1$ norm. This is why, even if we will present in the following results by trying to generalize for as many couples $(i,j)$ as possible, in practice only the $\hat{p}_n^{1,j}$, for all $j\geq 1$, interest us. 
 
\begin{Proposition}[Existence and uniqueness] \label{thm:existence}
The estimator $\hat{p}_n^{i,j}$ defined by \eqref{optbarycenternormed} exists for all $i\geq 1, j\geq 1$. 

$\hat{p}_n^{i,j}$ is unique if and only if $i\neq 1$.

In the following $\hat{p}_n^{1,1}$ therefore refers to a set of probability measures. 
\end{Proposition}

\begin{Proof}
See annex \ref{proofbarycenterexistence}.
\end{Proof}

The next proposition shows that one of the solutions of \eqref{optbarycenternormed} can always be written as a barycenter between $p_n^{\text{emp}}$ and $p^{\text{expert}}$ when $i=j$. This property therefore provides us, in these cases, with an explicit expression of a solution of \eqref{optbarycenternormed} which was not otherwise trivial to obtain by a direct calculation looking for the saddle points of the Lagrangian (for example in the case $i=j=1$).  

\begin{Proposition}
\label{barycenterNormed}
Let $\hat{p}_n^{i,j}$ defined by \eqref{optbarycenternormed} then for all $i=j$, it exists $\tilde{p}\in\hat{p}_n^{i,j}$ such that $\exists \alpha_n \in[0,1]$:
\begin{equation}
    \tilde{p}=\alpha_n p^{\text{expert}} + (1-\alpha_n) p_n^{\text{emp}}
    \label{formebarycentrique}
\end{equation}
where $\alpha_n=\displaystyle\frac{\epsilon_n}{\lVert p_n^{\text{emp}}-p^{\text{expert}}\rVert_i}$ if $\epsilon_n\leq \lVert p_n^{\text{emp}}-p^{\text{expert}}\rVert_i$ and $\alpha_n=1$ otherwise.
\end{Proposition}

\begin{Proof}
See annex \ref{proofbarycentercomputation}.
\end{Proof}

In particular one of the elements of $\hat{p}_n^{1,1}$ can be written under the form of a barycenter. For the sake of simplicity, we will designate in the following by $\hat{p}_n^{1,1}$ the solution of \eqref{optbarycenternormed} for $i=j=1$ which can be written under the form \eqref{formebarycentrique} and no more the whole set of solutions.   
\begin{Remark}
Note that the proposition \ref{barycenterNormed} is not true when $i=1$ and $j\neq 1$. This is why we focus on $\hat{p}^{1,1}_n$ for the end of this section.
\end{Remark}

It is now a question of deriving a result proving that mixing experts and data as we do with $\hat{p}_n^{1,1}$ represents an interest rather than choosing binary one of the two models. For this reason, we show in the following proposition that with a high probability, our estimator $\hat{p}^{1,1}$ is always better than the best of the models within a constant.  

\begin{Theorem}\label{whatisbesteuclidean}
Let $\hat{p}_n^{1,1}$ defined by \eqref{optbarycenternormed}. Then we have with probability at least $1-\delta$:
\begin{equation}
    \lVert p^\star - \hat{p}_n^{1,1} \rVert_1 \leq 2 \min \{ \epsilon_n, \lVert p^\star - p^{\text{expert}} \rVert_1 \}
\end{equation}
\end{Theorem}

\begin{Proof}
See annex \ref{bestbothworldlp}.
\end{Proof}

\subsection{Barycenter using the Kullback-Leibler divergence  }

In this section we study the theoretical properties of the solution of equation \eqref{formulegenerale} in the particular case where the dissimilarity measure $\mathcal{L}$ is the Kullback-Leibler divergence.  

The Kullback-Leibler divergence between two discrete probability measure $p$ and $q$ is defined as:

\begin{equation*}
    \mathbb{KL}(p||q)=\displaystyle\sum_i p_i \log\left(\displaystyle\frac{p_i}{q_i}\right).
\end{equation*}

Let us recall that the Kullback-Leibler divergence is not a distance since it is not symmetric and does not satisfies the triangular inequality, it is however positive defined \cite{Cover:2006:EIT:1146355}. 

We define our estimator as : 
\begin{equation}\label{LeftKullbackprojection}
\hat{p}_n^L=\argmin_{p\in \mathcal{C}/ \mathbb{KL}(p_n^{\text{emp}} || p ) \leq \epsilon_n}  \mathbb{KL}(p^{\text{expert}} || p )
\end{equation}

where \begin{equation}\label{confidence2}
\epsilon_n:=\epsilon_n^{\delta}=\argmin_{l} \mathbb{P}[\mathbb{KL}( p_n^{\text{emp}}||p^{\star})\leq l]\geq 1-\delta.
\end{equation}

To calibrate $\epsilon_n$, we can use the concentration inequality obtained in \cite{mardia2018concentration}. More precisely we have : 

\begin{equation}
    \epsilon_n=\displaystyle\frac{1}{n}\left(-\log(\delta)+\log\underbrace{\left(3+3\displaystyle\sum_{i=1}^{K-2} \left(\sqrt{\displaystyle\frac{e^3 n}{2\pi i }}\right)^i\right)}_{=:G_n} \right).
    \label{epsilonKullback}
\end{equation}

In the following proposition, we show the existence and uniqueness of our estimator $\hat{p}_n^L$ and the fact that our estimator is a barycenter. However it does not seem possible this time, unlike the case of $\hat{p}_n^{1,1}$ of the equation \eqref{optbarycenternormed}, to obtain a closed form for $\hat{p}_n^L$. 

\begin{Proposition}
\label{barycenterKullback}
Let $\hat{p}_n^L$ defined by \eqref{LeftKullbackprojection} then $\hat{p}_n^L$ exists and is unique. Moreover $\hat{p}_n^L$ can be written under the following form : 

\begin{equation}
    \hat{p}_n^L=\dfrac{1}{1+\tilde{\lambda}}p^{\text{expert}}+\displaystyle\frac{\tilde{\lambda}}{1+\tilde{\lambda}}p_n^{\text{emp}}
    \label{interpolationKullback}
\end{equation}

where $\tilde{\lambda}$ is a non-negative real such that: 

\begin{equation}
    \tilde{\lambda} \geq \displaystyle\frac{\mathbb{KL}(p_n^{\text{emp}}||p^{\text{expert}})}{\epsilon_n}-1.
    \label{lambdatilde}
\end{equation}
\end{Proposition}

\begin{Proof}
See annex \ref{kullbackuniqueness}.
\end{Proof}

The following proposition is intended to be the analog of the proposition \ref{whatisbesteuclidean} when $\mathcal{L}$ is the Kullback-Leibler divergence. We prove that the centroid $\hat{p}_n^L$ is better than the experts (with high probability). On the other hand, we obtain that when $\mathbb{KL}(p_n^{\text{emp}}|||p^\star)>\mathbb{KL}(p^{\text{expert}}||p^\star)$, the $\hat{p}_n^L$ barycenter is better than the empirical distribution. To obtain guarantees when $\mathbb{KL}(p_n^{\text{emp}}||p^\star)\leq \mathbb{KL}(p^{\text{expert}}||p^\star)$ seems less obvious and requires control over the quantity $\mathbb{KL}(p_n^{\text{emp}}||p^{\text{expert}})$.    

\begin{Theorem}\label{thmbestworldkl}
Let $\hat{p}_n^L$ defined by \eqref{LeftKullbackprojection} then we have with probability at least $1-\delta$: 
\begin{equation}
    \mathbb{KL}(\hat{p}_n^L || p^\star)\leq \min\left\{ \mathbb{KL}(p^{\text{expert}}|| p^\star),\epsilon_n \left(L_n+1\right)\right\}
\end{equation}

where 

\begin{equation*}
    L_n=\displaystyle\frac{\mathbb{KL}(p^{\text{expert}}||p^\star)-\mathbb{KL}(p_n^{\text{emp}}||p^\star)}{\mathbb{KL}(p_n^{\text{emp}}||p^{\text{expert}})}.
\end{equation*}

\end{Theorem}

\begin{Proof}
See annex \ref{kullbackbestbothworld}.
\end{Proof}

\begin{Remark}
Note that $\mathbb{KL}(\hat{p}_n^L || p^\star)$ is infinite if $p^\text{expert}$ does not have the same support that $p^\star$. Nevertheless obtaining a result for $\mathbb{KL}(p^\star|| \hat{p}_n^L)$ would require to have a concentration on $\mathbb{KL}(p^\star|| \hat{p}_n^{\text{emp}})$ which we do not have. Note that $\mathbb{KL}(p^\star|| \hat{p}_n^{\text{emp}})$ is infinite until we have sampled at least one time all the elements of the support of $p^\star$. 
\end{Remark}

\subsection{Some numerical results }
\label{numericBarycenter}

For each experiment in this section, we generate a random distribution $p^\star$ that we try to estimate. To do this, we simulate some realizations of a uniform distribution and renormalize in order to sum up to $1$. 

We also generate four different distributions that will serve as a priori for the inference: $p^{\text{expert},i},\forall i \in \{1,2,3,4\}$. The first three priors are obtained by a maximum entropy procedure under constraint to respect marginals of $p^\star$ having undergone a modification. We added to the marginals of $p^\star$ a Gaussian noise of zero expectation and variance equal to $\sigma_1^2=0.1$, $\sigma_2^2=0.2$ and $\sigma_3^2=0.4$ respectively. The last priority $p^{\text{expert},4}$ is chosen equal to the distribution $p^\star$ (the experts provided us with the right distribution).

We then sequentially sample data from $p^\star$, i.e we generate patients, and update for each new data and each different a priori, the left centroid $\hat{p}_n^L$ (using an Uzawa algorithm), the barycenter $\hat{p}^{1,1}_n$, the empirical distribution $p_n^{\text{emp}}$ as well as the divergences $\mathbb{KL}(\hat{p}_n^L||p^\star)$ and $\mathbb{KL}(p_n^{\text{emp}}||p^\star)$ and the norms $\lVert \hat{p}^{1,1}_n - p^\star \rVert_1$ and $\lVert p_n^{\text{emp}} - p^\star\rVert_1$. 

\begin{figure}
   \begin{minipage}[c]{.475\linewidth}     \includegraphics[scale=0.4]{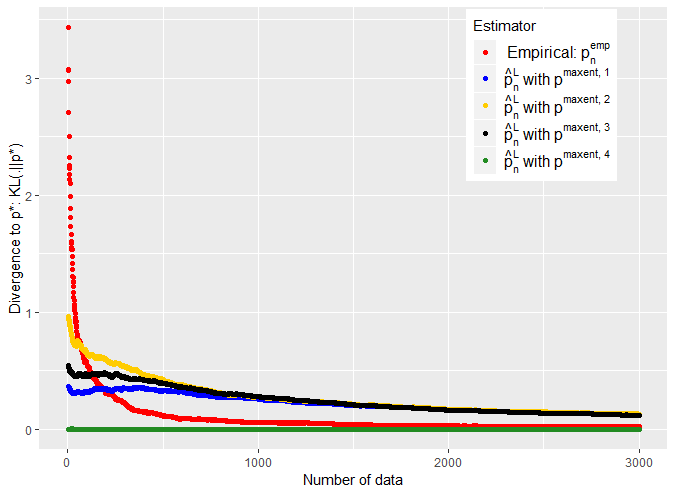}
   \caption{Evolution of the performance of $\hat{p}_n^L$ as a function of the available number of empirical data. $\epsilon_n$ defined by equation \eqref{epsilonKullbackbis}}
        \label{KullbackBarycenterConservateur}
   \end{minipage} \hfill
   \begin{minipage}[c]{.475\linewidth}     \includegraphics[scale=0.4]{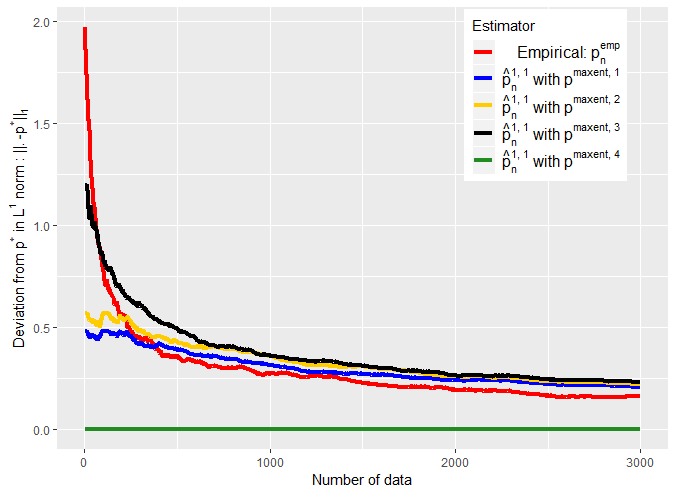}
      \caption{Evolution of the performance of $\hat{p}_n^{1,1}$ as a function of the available number of empirical data. $\epsilon_n$ defined by equation \eqref{epsilonEuclidianBarycenterbisbis}.} 
        \label{EuclidianBarycenterNonConservateurComplet}
   \end{minipage}
\end{figure}

\begin{figure}
   \begin{minipage}[c]{.475\linewidth}     \includegraphics[scale=0.4]{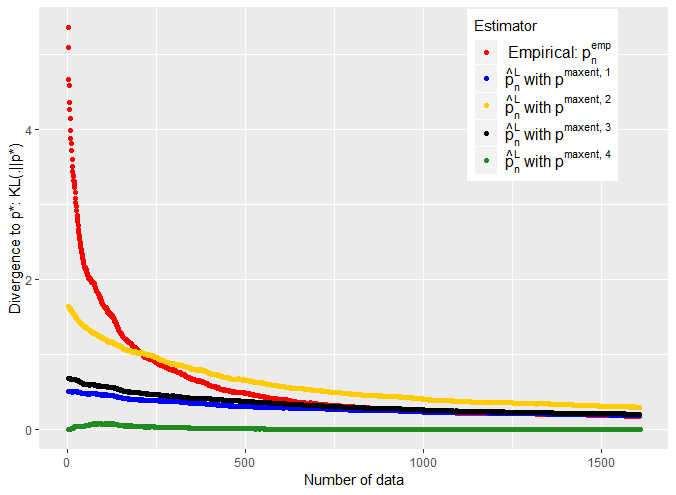}
        \caption{Evolution of the performance of $\hat{p}_n^L$ as a function of the available number of empirical data. $\epsilon_n$ defined by \eqref{epsilonKullbackBarycenterbisbis}. Number of symptom : $9$.}
        \label{KullbackBarycenterNonConservateurComplet}
   \end{minipage} \hfill
   \begin{minipage}[c]{.475\linewidth}     \includegraphics[scale=0.4]{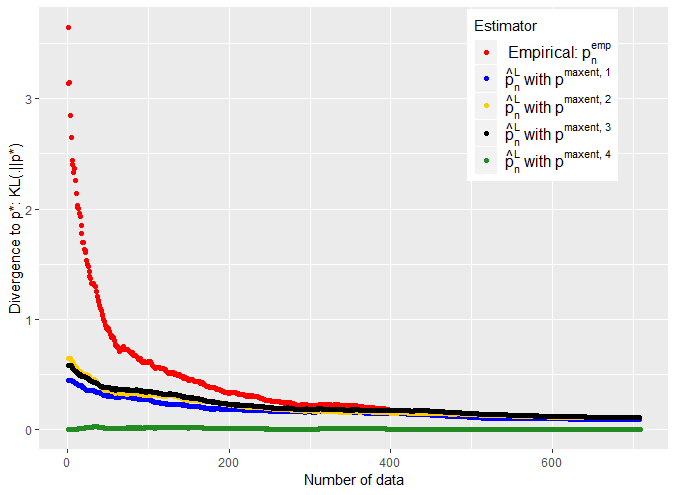}
        \caption{Evolution of the performance of $\hat{p}_n^L$ as a function of the available number of empirical data. $\epsilon_n$ defined by \eqref{epsilonKullbackBarycenterbisbis}. Number of symptom : $7$.}
        \label{KullbackBarycenterNonConservateurPasComplet}
   \end{minipage}
\end{figure}

\begin{figure}
\centering
   \includegraphics[scale=0.6]{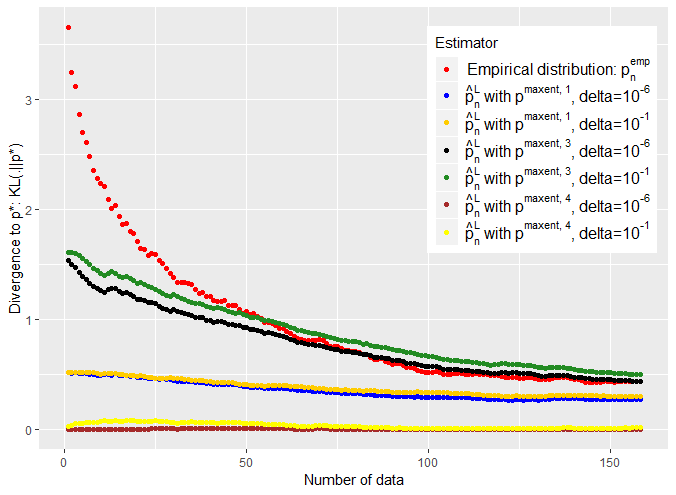}
        \caption{Evolution of the performance of $\hat{p}_n^L$ as a function of the available number of empirical data with different initial a priori and $\delta$. $\epsilon_n$ is defined by equation \eqref{epsilonKullbackbis}. Number of symptom : $7$.}
        \label{KullbackBarycenterDifferentDelta}
\end{figure}

The experiments of figures \ref{KullbackBarycenterConservateur}, \ref{EuclidianBarycenterNonConservateurComplet}, \ref{KullbackBarycenterNonConservateurPasComplet} were conducted on a case of a disease with $J=7$ typical symptoms and where there is therefore $K=2^7=128$ possible combinations. The experiments of figures \ref{KullbackBarycenterNonConservateurComplet} and \ref{KullbackBarycenterDifferentDelta} were conducted on a case of a disease with $9$ typical symptoms and where there is therefore $K=2^9=512$ possible combinations.

The only parameter we can control is the $\delta$ used to construct the confidence interval of the concentration of the empirical distribution around the true distribution. Let us recall, that for the case of the Kullback centroid of the equation \eqref{LeftKullbackprojection} we set : 

\begin{equation}
    \epsilon_n=\displaystyle\frac{1}{n}\left(-\log(\delta)+\log(G_n) \right)
    \label{epsilonKullbackbis}
\end{equation}

where $G_n$ is defined in equation \eqref{epsilonKullback}.

However, our first numerical experiments show that the choice of $\epsilon_n$ defined by the equation \eqref{epsilonKullbackbis} is a little too conservative: see figure \ref{KullbackBarycenterConservateur}. We need to converge $\epsilon_n$ faster towards $0$ without abandoning our a priori when it is good. 

Our experiments suggest taking a $\epsilon_n$ consistent with the proposed concentration in a conjecture of \cite{mardia2018concentration} for Kullback-Leibler divergence :

\begin{equation}
    \epsilon_n=\displaystyle\frac{-\log(\delta)+\frac{n}{2}\log\left(1+\frac{K-1}{n}\right)}{n}.
    \label{epsilonKullbackBarycenterbisbis}
\end{equation}

Note that we added a constant $\frac{1}{2}$ to the conjecture of \cite{mardia2018concentration}. As for the choice of $\delta$, this appears important mainly when $n$ is small, taking it sufficiently low avoids an overfitting situation when the number of data is still low without being harmful when $n$ is high. We took it equal to $10^{-6}$ in all our experiments.

The figure \ref{KullbackBarycenterDifferentDelta} shows that our approach is not very sensitive to the choice of $\delta$ which is an advantage compared to the penalized approach. 

The figures \ref{KullbackBarycenterNonConservateurPasComplet} and \ref{KullbackBarycenterNonConservateurComplet} show that such a choice for $\epsilon_n$ makes a good trade-off between expert and empirical data because we are able to take advantage of these two sources of information when the number of data is small (typically when $n<K$), but also to quickly abandon our a priori when it is bad (see the black curves) or to keep it when it is good (the green curves). Eventually the figures \ref{KullbackBarycenterNonConservateurPasComplet}, and \ref{KullbackBarycenterNonConservateurComplet} were performed on problems of $128$ and $512$ respectively and this choice of $\epsilon_n$ therefore appears relatively robust to changes in size. 

Concerning $\hat{p}_n^{1.1}$, we took, still following the conjectures of \cite{mardia2018concentration}:

\begin{equation}
    \epsilon_n=\sqrt{\displaystyle\frac{-\log(\delta)+\frac{n}{2}\log\left(1+\frac{K-1}{n}\right)}{n}}. 
    \label{epsilonEuclidianBarycenterbisbis}
\end{equation}

The figure \ref{EuclidianBarycenterNonConservateurComplet} shows the error made by our barycenter in norm $L^1$: $\hat{p}_n^{1,1}$ using such a $\epsilon_n$. We are again able to get rid of a bad a priori relatively quickly to follow the empirical (green curve) while keeping it if it is good (blue curve).   

Moreover we show with these experiments that there is an intermediate regime, when we do not have much data, where our estimator is \textit{strictly} better than the two models (experts and data alone). This is particularly visible when we used the $\epsilon_n$ of the conjecture of \cite{mardia2018concentration}, see figure \ref{KullbackBarycenterNonConservateurComplet} and \ref{KullbackBarycenterNonConservateurPasComplet}. It is then an empirical evidence that mixing these two heterogeneous sources of information, experts and empirical data, can be useful for statistical inference.   

Note that there is some limitations to these experiments. The way we simulate the distributions we are trying to estimate (the $p^\star$) produces quite specific distributions: close to the uniform and dense. If we simulate more sparse distributions, our a priori $p^{\text{expert}}$ built from the entropy maximum heuristic will be bad and the experiments will no longer be as interesting.

\section{Conclusion and perspectives}

In this work we have presented a way to combine expert knowledge, taking the form of marginal probabilities and rules, together with empirical data so as to estimate a given discrete probability distribution. This problem has emerged from our application where we aim to learn the probability distribution of the different combinations of symptoms given the disease. For this objective we have an initial a priori consisting of the marginal distributions coming from the medical literature and clinical data collected as the decision support tool is used.   

The particular form of the prior does not allow us to simply adopt a maximum a posteriori (MAP) approach. The absence of an obvious a priori to modelize the parameter's distribution naturally leads us to the idea of maximum entropy : if no model seems more plausible to us than another, then we will choose the least informative. 

This idea of maximum entropy brings us back to the works of the 80s' and 90 s' where researchers also aimed to build a symptom checker using the marginals. In our work we go further by gradually integrating empirical data as the algorithm is used. 

We are interested in the intermediate regime where we do not have enough empirical data to do without experts but have enough to correct them if necessary. Our proposal is to construct our estimator as the distribution closest to the experts' initial a priori, in the sense of a given dissimilarity measure, that is consistent with the empirical data collected. 

We prove, both theoretically and empirically, that our barycenter estimator mixing the two sources of information is always more efficient than the best of the two models (clinical data or experts alone) within a constant. 

We have empirically illustrated the effectiveness of the proposed approach by giving an a priori of different quality and incrementally adding empirical data. We have shown that our estimator allows a bad  a priori to be abandoned relatively quickly when the inconsistency of the data collected with the initial a priori is observed. At the same time, this same mixture makes it possible to keep the initial a priori if it is good. Moreover we show with this experiment that, in the intermediate regime, our estimator can be \textit{strictly} better than the best of the two models (experts and data alone). It empirically confirms the idea that mixing these two heterogeneous sources of information can be profitable in statistical inference.   

Future work will concentrate on several refinements such as the addition of a kernel structure for the construction of the empirical distribution. Indeed it is possible that there are omissions of some symptoms in the data collected. Then a kernel approach that would consider states that only differ by some presences as closer would capture such a difficulty and makes a better use of empirical data. Other dissimilarity measures could also be investigated.

\appendix

\section{Proof of the theoretical results of our barycenter estimator in the $L^p$ spaces}\label{proofbarycenter}

\subsection{Existence and uniqueness}\label{proofbarycenterexistence}

\begin{Proof}[Proof of Proposition \ref{thm:existence}]
The existence of a solution of \eqref{optbarycenternormed} for all $i\geq 1$ and $j\geq 1$ is a consequence of the fact that the projection onto a finite dimension set always exists.

The uniqueness of a solution of \eqref{optbarycenternormed} for all $i\neq 1$ is due to the fact that we aim to minimize a strictly convex function under convex constraints. When $j=1$ the function that we aim to minimize is no longer strictly convex and some counter-examples can be exhibited. 

For example if $p^{\text{expert}}=(\frac{1}{4},\frac{1}{4},\frac{1}{4},\frac{1}{4})$, $p_n^{\text{emp}}=(\frac{1}{2},0,\frac{1}{2},0)$ et $\epsilon_n=\frac{9}{10}$. 

Note that $\lVert  p^{\text{expert}} - p_n^{\text{emp}}\rVert_1=1 >\frac{9}{10}$. Then using proposition \ref{barycenterNormed} we know that
\begin{align*}
   &\displaystyle\frac{\epsilon_n}{\lVert p_n^{\text{emp}}-p^{\text{expert}} \rVert_1} p^{\text{expert}} + \left(1-\frac{\epsilon_n}{\lVert p_n^{\text{emp}}-p^{\text{expert}} \rVert_1}\right)p_n^{\text{emp}}\\
   &=\left(\frac{11}{40},\frac{9}{40},\frac{11}{40},\frac{9}{40}\right)=:\hat{p}_n^{1,1} 
\end{align*} is solution. But $\tilde{p}=(\frac{10}{40},\frac{9}{40},\frac{12}{40},\frac{9}{40})$ is solution too. Indeed : $$\lVert \tilde{p}-p^{\text{expert}} \rVert_1=\frac{1}{10}=\lVert \hat{p}^{1,1}_n-p^{\text{expert}} \rVert_1$$
and:
$$\lVert\tilde{p}-p_n^{\text{emp}}\rVert_1=\frac{36}{40} = \frac{9}{10}.$$  
\end{Proof}

\subsection{Linear combination of expert and data : a barycenter }\label{proofbarycentercomputation}
\begin{Proof}[Proof of Proposition \ref{barycenterNormed}]
 Let $\tilde{p}\in \mathcal{C}$ be such that it exists $\alpha\in [0,1]$ where $\tilde{p}=\alpha p^{\text{expert}}+(1-\alpha)p_n^{\text{emp}}$ and such that $\lVert \tilde{p} - p_n^{\text{emp}}\rVert_i=\epsilon_n$.
We then have :

\begin{equation*}
    \lVert \tilde{p} - p_n^{\text{emp}}\rVert_i=\alpha \lVert p_n^{\text{emp}}-p^{\text{expert}}\rVert_i=\epsilon_n
\end{equation*}

and then 

\begin{equation*}
    \alpha=\displaystyle\frac{\epsilon_n}{\lVert p_n^{\text{emp}}-p^{\text{expert}}\rVert_i}.
\end{equation*}

Moreover note that we have the following equality since $\tilde{p}$ can be written under the form of a barycenter : 

\begin{equation*}
    \lVert \tilde{p} -p^{\text{expert}}\rVert_i + \underbrace{\lVert \tilde{p} - p_n^{\text{emp}}\rVert_i}_{=\epsilon_n}=\lVert p_n^{\text{emp}} - p^{\text{expert}} \rVert_i
\end{equation*}

Let us make a reasoning by reductio ad absurdum. Let us assume that there exist $p'\in \mathcal{C}$ such that $\lVert p_n^{\text{emp}}-p'\rVert_i\leq \epsilon_n$ et $\lVert p^{\text{expert}}-p'\rVert_i<  \lVert p^{\text{expert}}-\tilde{p}\rVert_i$.

We would then have: 

\begin{align*}
    \lVert p_n^{\text{emp}}-p^{\text{expert}}\rVert_i &\leq \lVert p'-p^{\text{expert}}\rVert_i + \lVert p' - p_n^{\text{emp}}\rVert_i \\
    &< \lVert \tilde{p}-p^{\text{expert}}\rVert_i + \lVert p' - p_n^{\text{emp}}\rVert_i \\
    &= \lVert p_n^{\text{emp}}-p^{\text{expert}}\rVert_i -\epsilon_n + \lVert p' - p_n^{\text{emp}}\rVert_i \\
    &\leq  \lVert p_n^{\text{emp}}-p^{\text{expert}}\rVert_i 
\end{align*}

which leads to the desired contradiction.  

\end{Proof}

\subsection{Our barycenter estimator is more efficient than the best of the two models, expert or data, within a constant}\label{bestbothworldlp}
\begin{Proof}[Proof of Theorem \ref{whatisbesteuclidean}]
 A simple application of the triangular inequality gives us :

\begin{equation*}
     \lVert p^\star - \hat{p}_n^{1,1} \rVert_1 \leq \lVert p^\star - p_n^{\text{emp}} \rVert_1 + \lVert p_n^{\text{emp}} - \hat{p}_n^{1,1} \rVert_1
\end{equation*}

However $\lVert p_n^{\text{emp}} - \hat{p}_n^{1,1} \rVert_1\leq \epsilon_n$ by construction and we have $ \lVert p^\star - p_n^{\text{emp}} \rVert_1\leq \epsilon_n$ with probability at least $1-\delta$. 

In addition to that : 

\begin{equation*}
     \lVert p^\star - \hat{p}_n^{1,1} \rVert_1 \leq \lVert p^\star - p^{\text{expert}} \rVert_1 + \lVert p^{\text{expert}} - \hat{p}_n^{1,1} \rVert_1
\end{equation*}

However using the definition of $\hat{p}_n^{1,1}$ and assuming $\lVert p^\star - p_n^{\text{emp}} \rVert_1\leq \epsilon_n$ then:

\begin{equation*}
    \lVert p^{\text{expert}} - \hat{p}_n^{1,1} \rVert_1\leq  \lVert p^{\text{expert}}-p^\star \rVert_1.
\end{equation*}

We can conclude that if $\lVert p^\star - p_n^{\text{emp}} \rVert_1\leq \epsilon_n$, which happens with probability at least $1-\delta$, then :

\begin{equation}
    \lVert p^\star - \hat{p}_n^{1,1} \rVert_1 \leq 2\min \{ \epsilon_n , \lVert p^\star - p^{\text{expert}} \rVert_1 \}
\end{equation}

\end{Proof}

\section{Proof of the theoretical results of our barycenter estimator with the $\mathbb{KL}$ divergence}\label{proofbarycenterKL}

\subsection{Existence and uniqueness. Formula as a linear combination of experts and empirical data}\label{kullbackuniqueness}

\begin{Proof}[Proof of Proposition \ref{barycenterKullback}]
The existence and uniqueness of $\hat{p}_n^L$ is a consequence of the fact that $\mathcal{T}=\{p/ p\in \mathcal{C} \hspace{0.1cm}\text{and} \hspace{0.1cm} \mathbb{KL}(p_n^{\text{emp}}||p)\leq \epsilon_n\}$ is a convex set. Indeed let $p, q\in \mathcal{T}$, $\alpha\in[0,1]$ then using the classical log-sum inequality we have: 

\begin{equation*}
    \mathbb{KL}(p_n^{\text{emp}}||\alpha p +(1-\alpha) q) \leq \alpha \mathbb{KL}(p_n^{\text{emp}}|| p) + (1-\alpha) \mathbb{KL}(p_n^{\text{emp}}||q)\leq \epsilon_n. 
\end{equation*}

The Lagrangian associated to the optimization problem \eqref{LeftKullbackprojection} can be written as : 

\begin{align*}
    L(p,\lambda,\mu) &= \displaystyle\sum_{i} p_i^{\text{expert}} \log\left(\displaystyle\frac{p_i^{\text{expert}}}{p_i}\right) \\ &+\lambda \left(\displaystyle\sum_{i} p_i^{\text{emp}} \log\left(\displaystyle\frac{p_i^{\text{emp}}}{p_i}\right)-\epsilon_n\right)\\ &+\mu \left(\displaystyle\sum_{i} p_i-1 \right)
\end{align*}

Deriving for all $i \in[1,K]$:

\begin{equation*}
\displaystyle\frac{\partial L(p,\lambda,\mu)}{\partial p_i}=-\displaystyle\frac{p_i^{\text{expert}}}{p_i}-\lambda\displaystyle\frac{p_i^{\text{emp}}}{p_i}+\mu 
\end{equation*}

Equating this last expression to $0$ and using the fact that the probability measures sums to $1$ we find : $\mu=\lambda+1$. Then we have for all $i \in[1,K]$: 
\begin{equation*}
      p_i=\displaystyle\frac{1}{1+\lambda}p_i^{\text{expert}}+\displaystyle\frac{\lambda}{1+\lambda}p_i^{\text{emp}}
\end{equation*}

We know that $\hat{p}_n^L$ exists and is unique and using the Kuhn-Tucker theorem (whose assumptions we satisfy since we minimize a convex function under convex inequality constraints) we know that the minimum of the optimization problem \eqref{LeftKullbackprojection} is reached for the saddle-point of the Lagrangian : $(\tilde{\lambda},\tilde{p})=(\tilde{\lambda},\hat{p}_n^L)$. We can then write :

\begin{equation}
      \hat{p}_n^L=\displaystyle\frac{1}{1+\tilde{\lambda}}p^{\text{expert}}+\displaystyle\frac{\tilde{\lambda}}{1+\tilde{\lambda}}p_n^{\text{emp}}
      \label{interpolationKullbackbis}
\end{equation}

 We could not obtain a closed form for $\tilde{\lambda}$ unlike the case of $\hat{p}^{i,i}$. however we know that by construction  $\mathbb{KL}(p_n^{\text{emp}}||\hat{p}_n)\leq \epsilon_n.$ 

Moreover using the log-sum inequality and our interpolation formula \eqref{interpolationKullbackbis} we have: 
\begin{align*}
    \mathbb{KL}(p_n^{\text{emp}}||\hat{p}_n)&= \mathbb{KL}\left(p_n^{\text{emp}}||\displaystyle\frac{1}{1+\tilde{\lambda}}p^{\text{expert}}+\displaystyle\frac{\tilde{\lambda}}{1+\tilde{\lambda}}p_n^{\text{emp}}\right)\\ &\leq \displaystyle\frac{1}{1+\tilde{\lambda}}\mathbb{KL}(p_n^{\text{emp}}||p^{\text{expert}}). 
\end{align*}

We then have the following condition under $\tilde{\lambda}$ :

\begin{equation*}
    \displaystyle\frac{1}{1+\tilde{\lambda}}\mathbb{KL}(p_n^{\text{emp}}||p^{\text{expert}})\leq \epsilon_n \Leftrightarrow \tilde{\lambda} \geq \displaystyle\frac{\mathbb{KL}(p_n^{\text{emp}}||p^{\text{expert}})}{\epsilon_n}-1
\end{equation*}

\end{Proof}

\subsection{Our barycenter is more efficient than the best of the two models, expert or data, within a constant }\label{kullbackbestbothworld}

\begin{Proof}[Proof of Theorem \ref{thmbestworldkl}]
Using the proposition \ref{barycenterKullback} we have : 

\begin{align*}
    \mathbb{KL}(\hat{p}_n^L||p^\star)&=\mathbb{KL}\left(\displaystyle\frac{1}{1+\tilde{\lambda}}p^{\text{expert}}+\displaystyle\frac{\tilde{\lambda}}{1+\tilde{\lambda}}p_n^{\text{emp}}||p^\star\right)\\
    &\leq \displaystyle\frac{1}{1+\tilde{\lambda}} \mathbb{KL}(p^{\text{expert}}||p^\star)
    +\displaystyle\frac{\tilde{\lambda}}{1+\tilde{\lambda}}\mathbb{KL}(p_n^{\text{emp}}||p^\star)\\
    &= \displaystyle\frac{1}{1+\tilde{\lambda}}\left(\mathbb{KL}(p^{\text{expert}}||p^\star)
    -\mathbb{KL}(p_n^{\text{emp}}||p^\star)\right)\\ &+\mathbb{KL}(p_n^{\text{emp}}||p^\star)\\
    &\leq \epsilon_n\left(\displaystyle\frac{\mathbb{KL}(p^{\text{expert}}||p^\star)-\mathbb{KL}(p_n^{\text{emp}}||p^\star)}{\mathbb{KL}(p_n^{\text{emp}}||p^{\text{expert}})}\right)\\
   &+\mathbb{KL}(p_n^{\text{emp}}||p^\star)
   \end{align*}

where we used the available inequality to $\tilde{\lambda}$ (the proposition \ref{barycenterKullback}) in the last inequality and the desired result is obtained by assuming that $\mathbb{KL}(p_n^{\text{emp}}||p^\star)\leq \epsilon_n$ which happens with probability at least $1-\delta$.

In addition, note that :

\begin{align*}
   \epsilon_n\left(\displaystyle\frac{\mathbb{KL}(p^{\text{expert}}||p^\star)-\mathbb{KL}(p_n^{\text{emp}}||p^\star)}{\mathbb{KL}(p_n^{\text{emp}}||p^{\text{expert}})}\right)&+\mathbb{KL}(p_n^{\text{emp}}||p^\star) \\
    &\leq \mathbb{KL}(p^{\text{expert}}||p^\star) \\
   \Leftrightarrow \epsilon_n \leq \mathbb{KL}(p_n^{\text{emp}}||p^{\text{expert}})
\end{align*}

However, if $\epsilon_n \geq \mathbb{KL}(p_n^{\text{emp}}||p^{\text{expert}})$ we have by construction that $\hat{p}_n^L=p^{\text{expert}}$ and therefore $ \mathbb{KL}(\hat{p}_n^L||p^\star)= \mathbb{KL}(p^{\text{expert}}||p^\star).$

We can conclude from all this that :

\begin{equation*}
     \mathbb{KL}(\hat{p}_n^L||p^\star)\leq \mathbb{KL}(p^{\text{expert}}||p^\star).
\end{equation*}

\end{Proof}

\section*{Acknowledgements}
R\'emi Besson thanks Fr\'edéric Log\'e-Munerel for fruitful discussions about this work.

This work is supported by a public grant overseen by the French National research Agency (ANR) as
part of the "Investissement d’Avenir" program, through the "IDI 2017" project funded by the IDEX
Paris-Saclay, ANR-11-IDEX-0003-02.

\bibliographystyle{plain}

\end{document}